\documentclass[conference]{IEEEtran}
\IEEEoverridecommandlockouts
\usepackage{cite}
\usepackage{amsmath,amssymb,amsfonts}
\usepackage{algorithmic}
\usepackage{graphicx}
\usepackage{hyperref}
\usepackage{textcomp}
\usepackage{xcolor}
\def\BibTeX{{\rm B\kern-.05em{\sc i\kern-.025em b}\kern-.08em
    T\kern-.1667em\lower.7ex\hbox{E}\kern-.125emX}}
\begin{document}

\title{Interpretable Deep Learning for Automatic Diagnosis of 12-lead Electrocardiogram\\

}

\DeclareRobustCommand*{\IEEEauthorrefmark}[1]{%
  \raisebox{0pt}[0pt][0pt]{\textsuperscript{\footnotesize #1}}%
}

\author{\IEEEauthorblockN{Dongdong Zhang\IEEEauthorrefmark{1,2},
Xiaohui Yuan\IEEEauthorrefmark{2},
Ping Zhang\IEEEauthorrefmark{1,3,*}}
\IEEEauthorblockA{\IEEEauthorrefmark{1}Department of Biomedical Informatics,
The Ohio State University, Columbus, USA}
\IEEEauthorblockA{\IEEEauthorrefmark{2}School of Computer Science and Technology, Wuhan University of Technology, Wuhan, China}
\IEEEauthorblockA{\IEEEauthorrefmark{3}Department of Computer Science and Engineering, The Ohio State University, Columbus, USA}
\IEEEauthorblockA{\IEEEauthorrefmark{*}Corresponding Author: zhang.10631@osu.edu}
}

\maketitle

\begin{abstract}
Electrocardiogram (ECG) is a widely used reliable, non-invasive approach for cardiovascular disease diagnosis. With the rapid growth of ECG examinations and the insufficiency of cardiologists, accurate and automatic diagnosis of ECG signals has become a hot research topic. Deep learning methods have demonstrated promising results in predictive healthcare tasks. In this paper, we developed a deep neural network for multi-label classification of cardiac arrhythmias in 12-lead ECG recordings. Experiments on a public 12-lead ECG dataset showed the effectiveness of our method. The proposed model achieved an average area under the receiver operating characteristic curve (AUC) of 0.970 and an average F1 score of 0.813. The deep model showed superior performance than 4 machine learning methods learned from extracted expert features. Besides, the deep models trained on single-lead ECGs produce lower performance than using all 12 leads simultaneously. The best-performing leads are lead I, aVR, and V5 among 12 leads. Finally, we employed the SHapley Additive exPlanations (SHAP) method to interpret the model's behavior at both patient level and population level. Our code is freely available at \url{https://github.com/onlyzdd/ecg-diagnosis}.
\end{abstract}


\section{Introduction}
Cardiovascular diseases (CVDs) are the leading cause of death and produce immense health and economic burdens in the United States and globally \cite{virani2020heart}. The electrocardiogram (ECG) is a simple, reliable, and non-invasive approach for monitoring patients' heart activity and diagnosing cardiac arrhythmias. A standard ECG has 12 leads including 6 limb leads (I, II, III, aVR, aVL, aVF) and 6 chest leads (V1, V2, V3, V4, V5, V6) recorded from electrodes on the body surface. Accurately interpreting the ECG for a patient with concurrent cardiac arrhythmias is challenging even for an experienced cardiologist and incorrectly interpreted ECGs might result in inappropriate clinical decisions or lead to adverse outcomes \cite{bogun2004misdiagnosis}.

An estimated 300 million ECGs are recorded worldwide annually \cite{holst1999confident} and keeps growing. Computer-aided interpretation of ECGs has become more important, especially in low-income and middle-income countries where experienced cardiologists are scarce \cite{world2014global}. Therefore, accurate and automatic diagnosis of ECG signals has become a hot research interest. In past decades, automatic diagnosis of ECGs has been widely investigated with the availability of large open-source ECG datasets such as MIT-BIH Arrhythmia database \cite{moody2001impact}, 2017 Physionet Challenge/CinC dataset \cite{clifford2017af}, 2018 China Physiological Signal Challenge dataset (CPSC2018) \cite{liu2018open}, PTB-XL database \cite{wagner2020ptb}.

In this study, we developed a deep neural network based on 1D convolutional neural networks for automatic multi-label classification of cardiac arrhythmias in 12-lead ECG recordings, and the model achieved comparable state-of-the-art performance (average F1 score is 0.813) on the CPSC2018 dataset. We also conducted experiments on single-lead ECGs and showed the performance of every single lead. In addition, we applied the SHapley Additive exPlanations (SHAP) method \cite{lundberg2017unified} to interpret the model's predictions at both patient level and population level.

To summarize, the contributions of our work are:

\begin{itemize}
    \item We developed a deep neural network for automatic diagnosis of cardiac arrhythmias and the results on a 12-lead ECG dataset showed its effectiveness.
    \item We compared the performance of the deep learning model with 4 baseline classifiers using expert features. The result showed the deep learning model outperformed all baseline classifiers.
    \item We conducted experiments on single-lead ECGs and the results suggested the F1 score, averaged across diagnostic classes, of the deep model trained on single-lead ECGs is 4.4\% to 11.8\% lower than using all 12 leads, and the top-performing single leads are lead I, aVR, and V5.
    \item To better understand the model's behavior, we employed the SHAP method to enhance clinical interpretability at both patient level and population level.
\end{itemize}

The rest of the paper is organized as follows. In Section II, we summarize related works on the automatic diagnosis of ECG. A detailed description of the 12-lead ECG dataset, deep neural network architecture, evaluation metrics, and model interpretability are provided in Section III. We present the experiment results and model interpretability in Section IV. Finally, we conclude the paper in Section V.

\section{Related Works}
\begin{figure}[t!]
    \centering
    \includegraphics[width=0.5\textwidth]{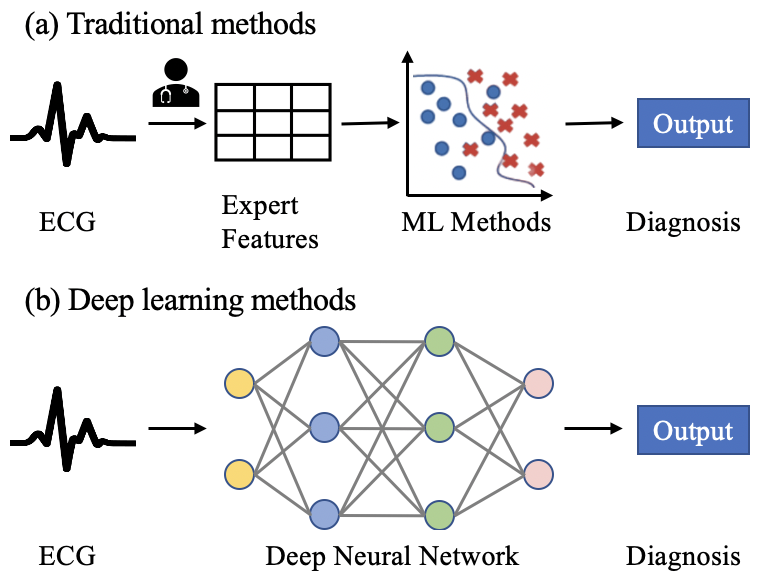}
    \caption{Comparison of existing models for automatic diagnosis of ECG abnormalities: (a) two-stage traditional methods using feature engineering; (b) end-to-end deep learning methods.}
    \label{fig:ml-deep}
\end{figure}

Existing models for automatic diagnosis of ECG abnormalities can be classified into two categories: traditional methods and deep learning methods. The comparison between traditional methods and deep learning methods are demonstrated in Fig. \ref{fig:ml-deep}. Traditional methods based on machine learning (ML) algorithms are two-stage, these methods require experts to engineer useful features or extract features using signal processing techniques first and then use these features to build machine learning classifiers \cite{jambukia2015classification, macfarlane2005university}. The University of Glasgow (Uni-G) ECG analysis program applied rule-based criteria on signal processing features and medical features for the diagnosis of ECGs \cite{macfarlane2005university}. The use of wavelet coefficients for the classification of ECGs has been investigated in \cite{de2000using}. Detta et al. developed a feature-oriented method with a two-layer cascaded binary classifier and achieved the best performance in the 2017 Physionet/CinC Challenge for atrial fibrillation classification from single-lead ECGs \cite{datta2017identifying}.

However, traditional methods are limited by data quality and domain knowledge. Additional effort is required to extract expert features. The second approach is using end-to-end deep learning techniques that don't require an explicit feature extraction. Deep learning methods have made great progress in many areas \cite{lecun2015deep} such as computer vision, speech recognition, and natural language processing since 2012. Many studies have also demonstrated promising results of deep learning in the healthcare domain such as complex diagnostics spanning dermatology, radiology, ophthalmology, and pathology \cite{esteva2019guide}. Recently, deep learning models have been applied to ECG data for various tasks including disease detection, annotation or localization, sleep staging, biometric human identification, denoising, and so on \cite{hong2020opportunities}. Deep neural networks have shown initial success in cardiac diagnosis from single-lead or multi-lead ECGs \cite{datta2017identifying, hannun2019cardiologist, he2019automatic, zhu2020automatic, chen2020detection}. A deep learning model trained on a large single-lead ECG dataset with 91,232 ECG recordings shows superior performance than cardiologists for diagnosing 12 rhythm classes \cite{hannun2019cardiologist}. Ullah et al. transformed the 1D ECG time series into a 2D spectral image through short-time Fourier transform and trained a deep learning model to classify cardiac arrhythmias \cite{ullah2020classification}. 12-lead ECGs are the standard techniques in realistic clinical settings and can provide more valuable information compared to single-lead ECGs. Chen et al. proposed an artificial neural network that combined convolutional neural networks (CNNs), recurrent neural networks (RNNs), and attention mechanism for cardiac arrhythmias detection and won first place in the 2018 China Physiological Signal Challenge \cite{chen2020detection}. Zhu et al. applied a deep learning algorithm to 12-lead ECGs to diagnosis 20 types of cardiac abnormalities, and the model performance exceeded physicians trained in ECG interpretation \cite{zhu2020automatic}. Besides, some studies \cite{hong2017encase, liu2018automatic} showed that the performance of neural networks can be significantly improved by incorporating expert features. Despite the promising performance of deep learning models on cardiac arrhythmias diagnosis, deep learning models usually operate as black boxes, and understanding the model's behavior on making decisions is important and challenging.

\section{Methods}

\subsection{12-lead ECG Database}
\noindent \textbf{CPSC2018 Database} The 1st China Physiological Signal Challenge (CPSC) 2018 \cite{liu2018open} hosted during the 7th International Conference on Biomedical Engineering and Biotechnology released a freely large multi-label 12-lead ECG database collected from 11 hospitals in China. This database comprises 6877 12-lead ECGs lasting between 6 s and 60 s at a sampling rate of 500 Hz. These ECGs are labeled with 9 diagnostic classes: normal sinus rhythm (SNR), atrial fibrillation (AF), first-degree atrioventricular block (IVAB), left bundle branch block (LBBB), right bundle branch block (RBBB), premature atrial contraction (PAC), premature ventricular contraction (PVC), ST-segment depression (STD), ST-segment elevation (STE). Patient characteristics and diagnosis class prevalence of the CPSC2018 dataset are shown in Table \ref{tab:statistics}. As shown in Table. \ref{tab:statistics}, data imbalance and insufficiency problem is severe for cardiac arrhythmias diagnosis.

\begin{table}[t!]
\caption{Patient characteristics and diagnostic class prevalence on the CPSC2018 dataset.}
\begin{center}
\begin{tabular}{|l|c|c|c|c|}
\hline
\textbf{Class} & \textbf{Count (\%)} & \textbf{Male (\%)} & \textbf{Age} & \textbf{Duration} \\ \hline
SNR  &  918 (13.35) &  363 (39.54) & 41.56 (18.45) &  15.43 (7.64) \\
AF   & 1221 (17.75) &  692 (56.67) & 71.47 (12.53) &  15.07 (8.73) \\
IAVB &  722 (10.50) &  490 (67.87) & 66.97 (15.67) &  14.42 (7.08) \\
LBBB &  236 (3.43) &  117 (49.58) & 70.48 (12.55) &  15.10 (8.10) \\
RBBB & 1857 (27.00) & 1203 (64.78) & 62.84 (17.07) &  14.73 (9.00) \\
PAC  &  616 (8.96) &  328 (53.25) & 66.56 (17.71) & 19.30 (12.39) \\
PVC  &  700 (10.18) &  357 (51.00) & 58.37 (17.90) & 20.84 (15.39) \\
STD  &  869 (12.64) &  252 (29.00) & 54.61 (17.49) &  15.65 (9.79) \\
STE  &  220 (3.20) &  180 (81.82) & 52.32 (19.77) & 17.31 (10.74) \\
\hline
\multicolumn{5}{l}{Mean and standard deviation are reported for age and ECG duration (s).}
\end{tabular}
\label{tab:statistics}
\end{center}
\end{table}

\subsection{Network Architecture}
\begin{figure}[t!]
\centerline{\includegraphics[width=0.5\textwidth]{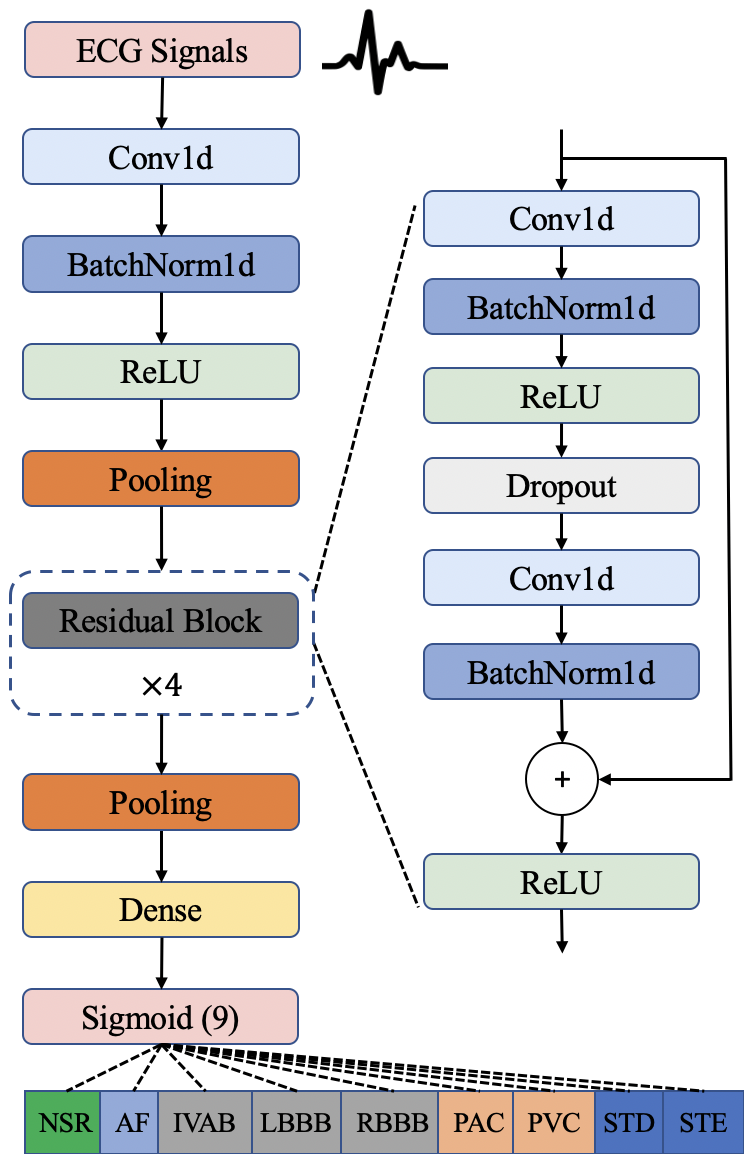}}
\caption{Deep neural network architecture for cardiac arrhythimas diagnosis. Our deep neural network accepts raw ECG inputs (12 leads, duration of 30 s, sampling rate of 500 Hz), utilizes 1D CNNs to extract deep features, and outputs the prediction results for 9 diagnostic classes.}
\label{fig:model}
\end{figure}

The overview of the proposed network architecture is illustrated in Fig. \ref{fig:model}. The proposed network is developed using 1D CNNs. Similar to the original residual neural network for image recognition with 2D CNNs \cite{he2016deep}, residual blocks with shortcut connections are utilized in our model to make the model training tractable. The model takes the raw ECG signals $x\in \mathbb{R}^{nsteps\times 12}$ (optimal value for $nsteps$ is 15000) as input and outputs a multi-label classification result $\hat{y} \in \mathbb{R}^{1\times 9}$. 

As shown in Fig. \ref{fig:model}, the network consists of 34 layers. 4 stacked residual blocks are used to extract deep features. Within each residual block, there are two 1D convolutional (Conv1d) layers, two batch normalization (BatchNorm1d) layers, 1 dropout (Dropout) layer, and two rectified linear unit (ReLU) activation layers. Conv1d layers are used to automatically extract features, BatchNorm1d layers to make the model faster and stable, ReLU layers to perform non-linear activation, Dropout layer to reduce overfitting. Conv1d layer with a kernel size of 1 and a max-pooling layer is used to match the dimensions and skip connections. The features extracted by stacked residual blocks are pooled using adaptive average-pooling and adaptive max-pooling. The pooling results are concatenated and then sent to the output layer with sigmoid as activation function to make predictions.

\subsection{Evaluation Metrics}
For each diagnostic class, we report Precision, Recall, F1 score (F1), area under the receiver operating characteristic curve (AUC), accuracy score (ACC). For class $i$, the metrics are calculated with the following equations:
\begin{IEEEeqnarray}{C}
    ACC_i = \cfrac{TP_i + TN_i}{TP_i + TN_i + FP_i + FN_i} \\
    Recall_i = \cfrac{TP_i}{TP_i + FN_i} \\
    Precision_i = \cfrac{TP_i}{TN_i + FP_i} \\
    F_{1i} = \cfrac{2 * Precision_i * Recall_i}{Precision_i + Recall_i}
\end{IEEEeqnarray}

where $TP_i$, $TN_i$, $FP_i$, and $FN_i$ represent the number of true positive samples, the number of true negative samples, the number of false positive samples, and the number of false negative samples for class $i$ respectively. Class $i$ can be one of the 9 classes: SNR, AF, IVAB, LBBB, RBBB, PAC, PVC, STD, and STE.

To better evaluate the performance of multi-label classification, we adopt average (AVG) score of each metric on 9 classes (1 normal and 8 abnormal). Average F1 score is used to select the best-performing model. And the final score is the average over classes:
\begin{IEEEeqnarray}{C}
    AUC = \cfrac{1}{9} \sum_{i=1}^9 AUC_i \\
    ACC = \cfrac{1}{9} \sum_{i=1}^9 ACC_i \\
    F_1 = \cfrac{1}{9} \sum_{i=1}^9 F_{1i}
\end{IEEEeqnarray}

\subsection{Interpretability}

\begin{figure*}[t!]
    \centering
    \includegraphics[width=\textwidth]{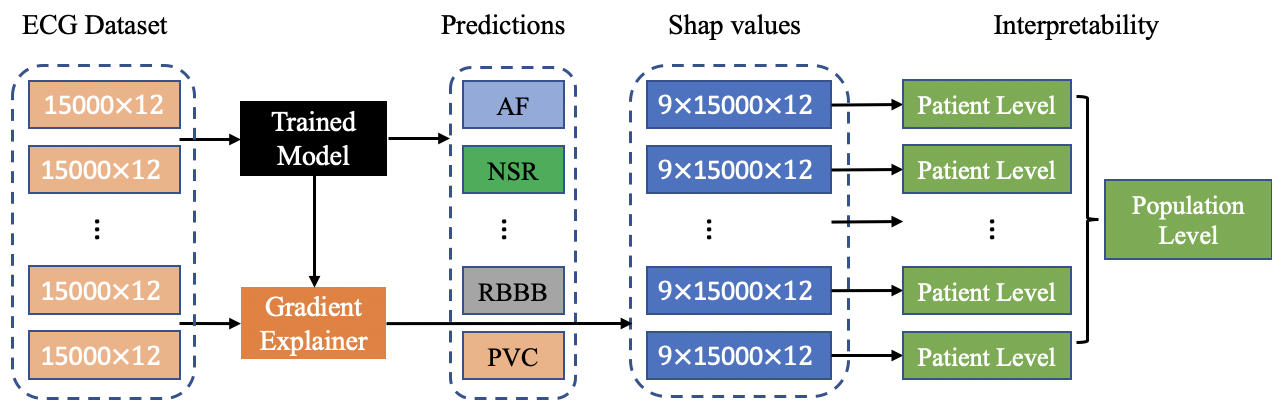}
    \caption{Interpretability of the deep learning model at both patient level and population level using shap values.}
    \label{fig:shap_intro}
\end{figure*}

Although deep learning models can achieve state-of-the-art performance in many predictive tasks, deep learning models are usually considered to be black boxes. Due to the multi-layer nonlinear structure, the decisions made by deep learning models are not traceable by humans. However, understanding the model's behavior when making predictions is as crucial as the accuracy of predictions in many applications, especially in clinical practice. To address this issue, we adopted the SHAP (SHapley Additive exPlanations) \cite{lundberg2017unified} method to interpret the model's predictions. SHAP is a game-theoretic approach to explain the model predictions and has been applied to tree-based algorithms to enhance clinical interpretability \cite{lundberg2020local, li2020time}. SHAP provides a unified way of interpreting predictions of any machine learning models. To be specific, SHAP assigns shap values, a unique additive feature importance measure, to each feature for a particular prediction. As shown in Fig. \ref{fig:shap_intro}, we applied the SHAP method to the trained deep learning model to interpret the model's behavior at both patient level and population level by utilizing a gradient explainer.

\subsubsection{Patient Level Interpretation}
Firstly, we focus on patient-level interpretation to understand why the model is making a certain prediction for 12-lead ECG inputs. Given an ECG input $x\in \mathbb{R}^{15000 \times 12}$, the model outputs a multi-label classification result $\hat{y} \in \mathbb{R}^{1\times 9}$. By applying the gradient explainer, a shap values matrix $sv\in \mathbb{R}^{9\times 15000 \times 12}$ is generated for each input where $sv_{i,j,k}$ represents the feature contribution of the corresponding ECG input $x_{j,k}$ towards the diagnostic class $i$. If $sv_{i,j,k}>0$, then $x_{j,k}$ contributes positively towards the diagnostic class $i$. For the top-predicted class $l={\rm argmax}(\hat{y})$, the submatrix $sv_l$ demonstrates why the deep learning model predicts $l$ given the ECG input $x$ and shows the contribution of features. 

\subsubsection{Population Level Interpretation}
While patient level interpretation explains the model's behavior on a specific ECG input, population level interpretation shows the contribution of ECG leads towards each kind of cardiac arrhythmias over the entire dataset. As shown in Fig. \ref{fig:shap_intro}, population level interpretation is the summarization of patient level interpretation. Given the population of $D$ patients and the shap values matrix $svs\in \mathbb{R}^{N \times 9\times 15000 \times 12}$, the contribution $c_{i,k}$ of lead $k$ for diagnostic class $i$ is defined as the sum of shap values:
\begin{IEEEeqnarray}{C}
    c_{i,k} = \sum_{n=1}^{D}\sum_{j=1}^{15000} svs_{d,i,j,k}
\end{IEEEeqnarray}

The normalized contribution rate $r_{i,k}$ of lead $k$ towards class $i$ is calculated as:
\begin{IEEEeqnarray}{C}
    r_{i,k} = \cfrac{c_{i,k}}{\sum_{i=1}^{12} c_{i,k}}
    \label{eq:ncr}
\end{IEEEeqnarray}

And the average contribution rate $\overline{r}_k$ of lead $k$ in 12-lead ECG model is:
\begin{IEEEeqnarray}{C}
    \overline{r}_k = \cfrac{1}{9}\sum_{i=1}^{9}{r_{i,k}}
    \label{eq:acr}
\end{IEEEeqnarray}

The normalized contribution rate $r_{i,k}$ shows which leads are playing an important role in diagnosing a particular cardiac arrhythmia $i$. The average contribution rate reflects the importance of each lead and implies possible feature interactions in the deep model.

\section{Results and Discussion}
\subsection{Experiment Setup}
\subsubsection{Data Preprocessing}
The CPSC2018 database comprises multi-label 12-lead ECGs with varying durations between 6 s and 60 s. As the deep neural network requires inputs to be of the same length. We preprocessed the dataset to make all inputs are of the same length $nsteps$. We tried different values for $nsteps$, and found that setting $nsteps$ to 15000 (duration of 30 s, sampling rate of 500 Hz) achieved the best performance. For ECGs with a duration of more than 30 s, they will be cropped and the last 30 s ECG data are kept. Otherwise, they will be padded to 30 s with zeros.

\subsubsection{Data Augmentation}
As shown in Table. \ref{tab:statistics}, data imbalance and insufficiency problem is severe for cardiac arrhythmias diagnosis. To address this problem, we applied scaling and shifting for data augmentation. Data augmentation can help reduce model overfitting and encourage robustness against adversarial examples.

\subsubsection{Training and Evaluation}
For model training and evaluation, we applied a 10-fold cross-validation approach. The CPSC2018 dataset was randomly divided into 10 folds. At each round, 8 folds out of 10 folds are used for training, 1 fold for validation, and 1 fold for testing. The optimal threshold of each class is selected to achieve the best F1 score on the validation dataset. Then the selected thresholds are applied to the test dataset to produce results. The reported results are the average on the test dataset of 10 rounds. The deep neural network is implemented using PyTorch \cite{paszke2019pytorch}. Adam optimizer \cite{jlb2015adam} is used as the optimization method and cross-entropy as the loss function to train the model. The optimal values for hyperparameters of the deep neural network are: the length of ECG input is set to 15000; the learning rate is 0.0001; the batch size is 32; the maximum number of epochs is 30; the kernel size of 1D CNNs is 15; the dropout rate of dropout layers is 0.2. Besides, our code is publicly available at \url{https://github.com/onlyzdd/ecg-diagnosis}.

\subsection{12-lead Model Performance}

Precision, Recall, F1 score, AUC, Accuracy of the model's prediction on each cardiac arrhythmia on the test dataset of 10 rounds are averaged and reported in Table. \ref{tab:result}. Overall, average AUC and Accuracy of the deep learning model both exceeded 0.95 and the average F1 score was 0.813 with an average precision of 0.821 and an average recall of 0.812. Among all cardiac arrhythmias, the deep model performed best on AF and RBBB classification with an F1 score of over 0.9. However, we also observed the F1 score of STE is low as 0.535 which may be due to the significant physician disagreement in diagnosing STE from ECGs \cite{mccabe2013physician}. 

\begin{table}[t!]
    \caption{12-lead model performance averaged on 10-fold tests.}
    \begin{center}
    \begin{tabular}{|c|c|c|c|c|c|}
    \hline
         & \textbf{Precision} & \textbf{Recall} & \textbf{F1}  &  \textbf{AUC}  & \textbf{Accuracy} \\ \hline
    SNR  &       0.814 &    0.800 &  0.805 &  0.974 &  0.948 \\ \hline
    AF   &       0.920 &    0.918 &  0.919 &  0.988 &  0.971 \\ \hline
    IAVB &       0.868 &    0.865 &  0.864 &  0.987 &  0.974 \\ \hline
    LBBB &       0.844 &    0.894 &  0.866 &  0.980 &  0.991 \\ \hline
    RBBB &       0.911 &    0.942 &  0.926 &  0.987 &  0.959 \\ \hline
    PAC  &       0.756 &    0.720 &  0.735 &  0.949 &  0.952 \\ \hline
    PVC  &       0.869 &    0.839 &  0.851 &  0.976 &  0.971 \\ \hline
    STD  &       0.808 &    0.826 &  0.814 &  0.971 &  0.953 \\ \hline
    STE  &       0.603 &    0.504 &  0.535 &  0.923 &  0.974 \\ \hline
    AVG  &       0.821 &    0.812 &  0.813 &  0.970 &  0.966 \\ \hline
    \end{tabular}
    \end{center}
    \label{tab:result}
\end{table}

\begin{figure}[t!]
\centerline{\includegraphics[width=0.5\textwidth]{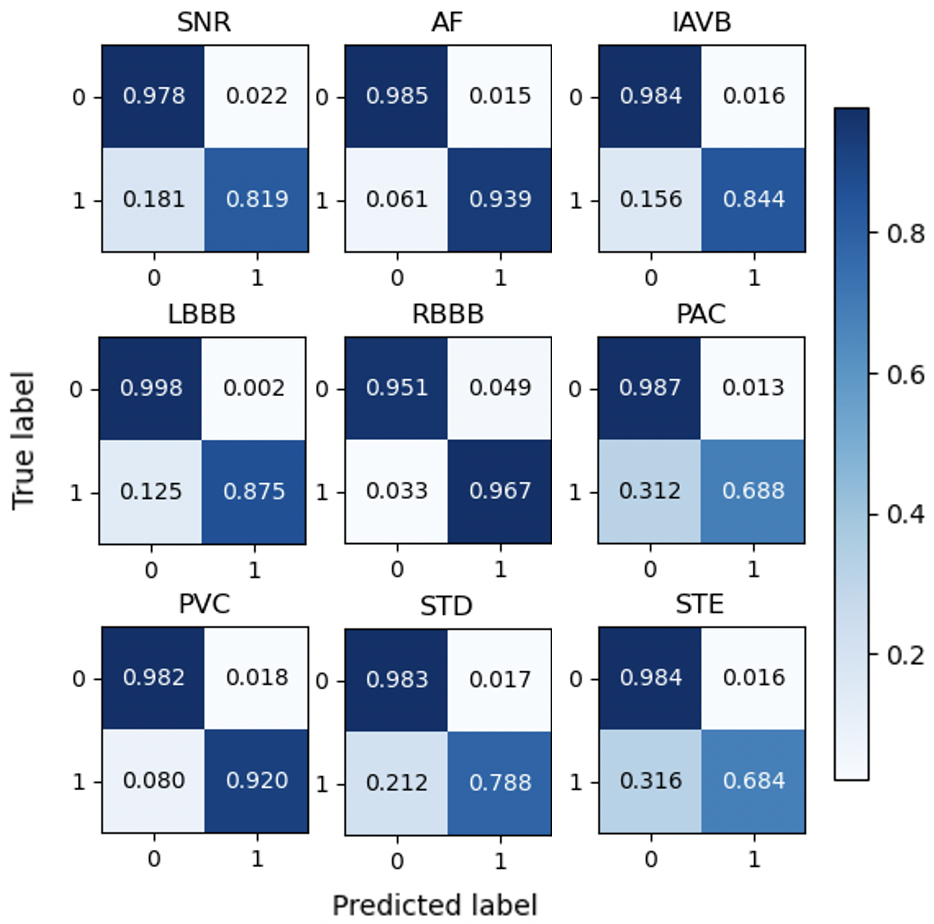}}
\caption{Multi-label confusion matrices of the best validation model predictions and ground truth.}
\label{fig:cm}
\end{figure}

Besides, we selected the best validation model of 10 rounds and used the confusion matrices calculated on the test dataset to illustrate why the model is working or not working on specific examples of cardiac arrhythmias. The confusion matrices are shown in Fig. \ref{fig:cm}. Low false negative rate and high true negative rate were observed for all 9 classes as shown in Fig. \ref{fig:cm}. For the diagnosis of AF, RBBB, and PVC, low false positive rate and false negative rate were observed. However, the confusion matrices showed that the model had trouble in classifying PAC, STD, and STE with a high false negative rate.

\subsection{Comparison with Traditional Methods}
Inspired by \cite{de2000using} and \cite{liu2018automatic}, we built several baseline models with extracted expert features. To be specific, we extracted 2 types of expert features: 1) statistical features (e.g., mean, standard deviation, variance, and percentile) of raw ECG input, 2) statistics and Shannon entropy of signal processing features extracted by applying discrete wavelet decomposition. Statistical features and signal processing features are concatenated and input to machine learning classifiers. For machine learning classifiers, we considered logistic regression (LR), random forest (RF), gradient boosting trees (GBT), and multi-layer perceptron (MLP).

\begin{figure*}[t!]
    \centering
    \includegraphics[width=\textwidth]{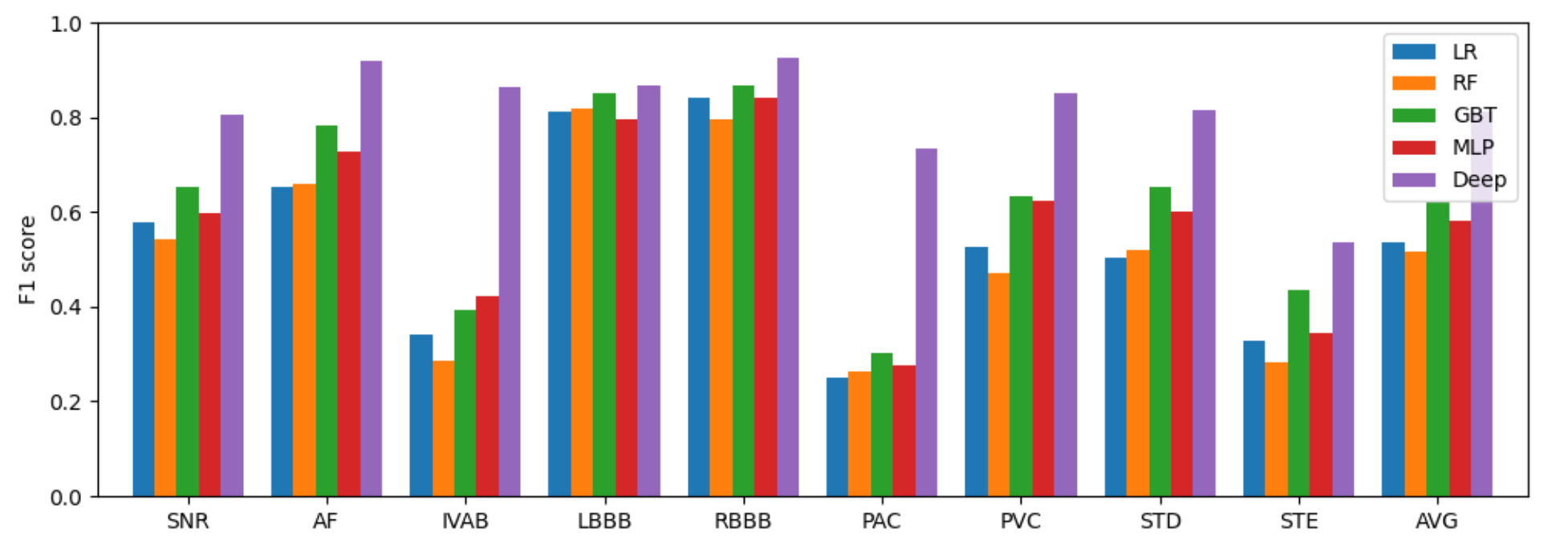}
    \caption{F1 score comparison of baseline models (LR, RF, GBT, MLP) using expert features and our end-to-end deep learning model (Deep).}
    \label{fig:comparison}
\end{figure*}

LR, RF, and MLP are implemented with scikit-learn toolkit \cite{pedregosa2011scikit}, and GBT is implemented using LightGBM \cite{ke2017lightgbm}. The comparison of model performance (F1 score) is shown in Fig. \ref{fig:comparison}. It's apparent that the deep learning model outperformed all baseline models with an average F1 score of 0.813. Among 4 baseline models, GBT achieved the best average F1 score of 0.619, while RF performed worst with an average F1 score of 0.515. As shown in Fig. \ref{fig:comparison}, the end-to-end deep learning model with deep features significantly improved the accuracy of IVAB and PAC classification. In addition, Fig. \ref{fig:comparison} showed the capability of extracted expert features in diagnosing AF, LBBB, and RBBB. However, these baseline models showed difficulty in classifying IVAB, PAC, and STE with low F1 scores. 

\subsection{Single-lead Model Performance}

\begin{table*}[t!]
\caption{Comparison of single-lead model performance measured by F1 score.}
\begin{center}
\begin{tabular}{|c|c|c|c|c|c|c|c|c|c|c|c|c|c|}
\hline
& \textbf{I} & \textbf{II} & \textbf{III} & \textbf{aVR} & \textbf{aVL} & \textbf{aVF} & \textbf{V1} & \textbf{V2} & \textbf{V3} & \textbf{V4} & \textbf{V5} & \textbf{V6} & \textbf{All} \\ \hline
SNR   &  0.705  &  0.682  &  0.602  &  0.712  &  0.604  &  0.663  &  0.657  &  0.694  &  0.710  &  0.717  &  0.731  &  0.721  &  \textbf{0.805} \\ \hline
AF    &  0.914  &  0.927  &  0.911  &  \textbf{0.929}  &  0.913  &  0.908  &  0.924  &  0.913  &  0.915  &  0.922  &  0.910  &  0.905  &  0.919 \\ \hline
IAVB  &  0.843  &  0.853  &  0.818  &  0.842  &  0.808  &  0.830  &  0.860  &  \textbf{0.866}  &  \textbf{0.866}  &  0.816  &  0.842  &  0.840  &  0.864 \\ \hline
LBBB  &  \textbf{0.897}  &  0.778  &  0.783  &  0.825  &  0.802  &  0.737  &  0.860  &  0.860  &  0.804  &  0.759  &  0.813  &  0.789  &  0.866 \\ \hline
RBBB  &  0.859  &  0.802  &  0.804  &  0.845  &  0.815  &  0.796  &  \textbf{0.940}  &  0.886  &  0.852  &  0.828  &  0.827  &  0.840  &  0.926 \\ \hline
PAC   &  0.723  &  \textbf{0.737}  &  0.709  &  0.688  &  0.698  &  0.719  &  0.730  &  0.689  &  0.692  &  0.680  &  0.715  &  0.702  &  0.735 \\ \hline
PVC   &  0.813  &  0.821  &  0.846  &  0.818  &  0.792  &  0.836  &  0.788  &  0.842  &  0.835  &  0.838  &  0.818  &  0.809  &  \textbf{0.851} \\ \hline
STD   &  0.695  &  0.790  &  0.627  &  0.793  &  0.573  &  0.711  &  0.615  &  0.652  &  0.702  &  0.753  &  0.781  &  0.757  &  \textbf{0.814} \\ \hline
STE   &  0.433  &  0.406  &  0.312  &  0.435  &  0.251  &  0.338  &  0.293  &  0.417  &  0.477  &  \textbf{0.552}  &  0.485  &  0.497  &  0.535 \\ \hline
AVG   &  0.765  &  0.755  &  0.712  &  0.765  &  0.695  &  0.726  &  0.741  &  0.758  &  0.762  &  0.763  &  0.769  &  0.762  &  \textbf{0.813} \\ \hline
\end{tabular}
\label{tab:single}
\end{center}
\end{table*}
We modified the input layer of the deep neural network and trained the model on single-lead ECG inputs $x\in \mathbb{R}^{15000 \times 1}$. Comparison of single-lead model performance measured by F1 score is summarized in Table. \ref{tab:single}. From Table. \ref{tab:single}, we observed: (1) In summary, the single-lead model showed inferior performance compared to using all 12 leads simultaneously. On average, the performance of deep learning model trained on single-lead ECGs dropped by 4.4\% to 11.8\% compared to using all 12 leads. (2) Among 12 leads, lead I, aVR, V5 are the top-performing single leads with an F1 score of more than 0.765, and lead aVL is shown to perform worst with an average F1 score of 0.695. (3) All single leads achieved good performance on AF classification with an F1 score of over 0.9. Lead II, aVR showed the comparable best performance in the diagnosis of AF. (4) The F1 score (0.94) on RBBB classification obtained using lead V1 is significantly higher than using any other leads which means V1 plays an important role in diagnosing RBBB. (5) The best predictive single lead for LBBB is lead I. (6) Lead I used by Apple Watch \cite{apple4ecg} and lead II favored by cardiologists for quick review also showed very good performance on average. (7) Interestingly, although 12-lead model achieved comparable or better performance than single-lead models for most diagnostic classes, lead I for LBBB and lead V1 for RBBB showed superior performance. We speculate that unexpected feature interactions may hurt the performance of the 12-lead model.

\subsection{Model Interpretability}
Model interpretability of deep neural networks has been a common challenge and limiting factor towards real-world applications. In addition to the promising performance achieved by our deep model in diagnosing cardiac arrhythmias, the SHAP method was used to explain model predictions. In this part, we demonstrated the model interpretability at both patient level and population level through visualizations.
\subsubsection{Patient Level Interpretation}
\begin{figure*}[t!]
    \centering
    \includegraphics[width=\textwidth]{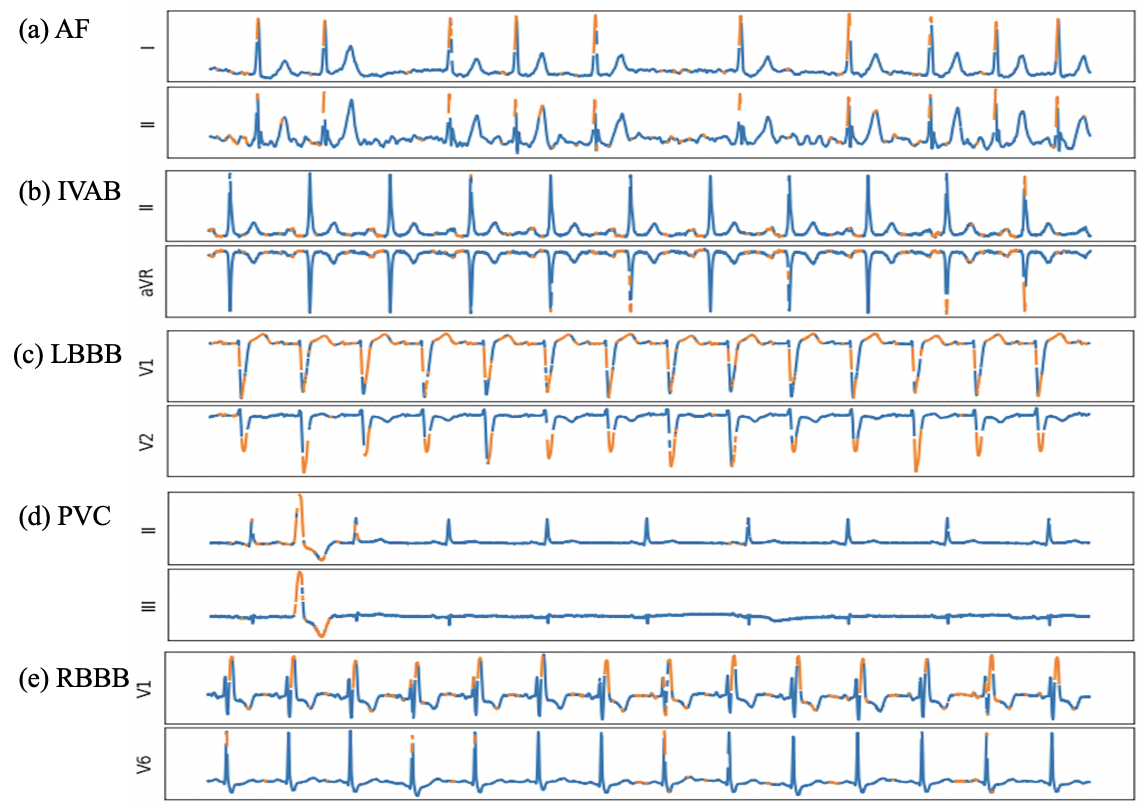}
    \caption{Explanation of the model's prediction results for several ECG instances from different patients. The features with high contribution (i.e., shap values) are highlighted in orange. Only the last 10 s of top 2 influential leads are displayed due to the limited space.}
    \label{fig:shap1}
\end{figure*}

For each ECG input with the top-predicted cardiac arrhythmia class $l={\rm argmax}(\hat{y})$, we visualized the shap values matrix $sv_l\in \mathbb{R}^{15000\times 12}$ along with the raw ECG input matrix $x\in \mathbb{R}^{15000\times 12}$. The explanations of the model's prediction results for several ECG instances from different patients are shown in Fig. \ref{fig:shap1}. For the ECG instance in Fig. \ref{fig:shap1}(a), the corresponding patient is diagnosed with AF. Highlighted areas in orange appear to include the P wave and R peak and may indicate a missing P wave and irregular RR intervals. This observation is consistent with the diagnostic criteria of AF \cite{gutierrez2011atrial}. In Fig. \ref{fig:shap1}(b) with IAVB, highlighted features mainly appear at PR intervals which are used for the diagnosis of IAVB \cite{barold2006first}. PVC happens in some sporadic periods in the ECGs, and only the period where PVC occurs is highlighted in Fig. \ref{fig:shap1}(d) which is reasonable. Fig. \ref{fig:shap1}(c) shows wide QRS complex in lead V1 for LBBB. Fig. \ref{fig:shap1}(e) shows rSR' pattern in lead V1 for RBBB. Observations from \ref{fig:shap1}(c) and Fig. \ref{fig:shap1}(e) are compatible with the corresponding diagnostic criteria for LBBB and RBBB \cite{alventosa2019right, goldberger2017clinical}.

\subsubsection{Population Level Interpretation}
\begin{figure}[t!]
    \centering
    \includegraphics[width=0.5\textwidth]{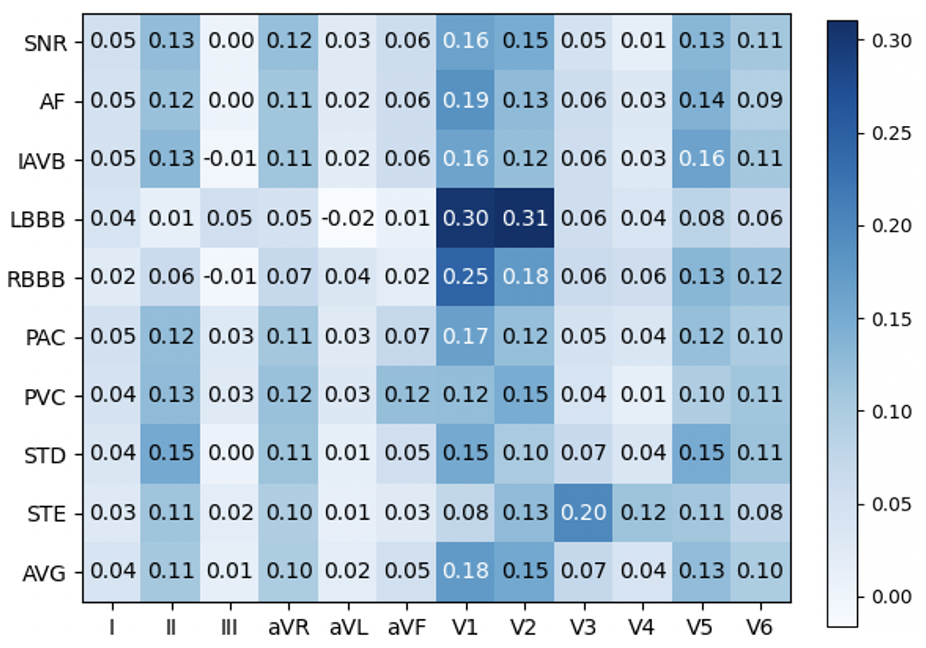}
    \caption{Population level interpretation by calculating the contribution rate of ECG leads towards diagnostic classes in the 12-lead deep model.}
    \label{fig:shap2}
\end{figure}

Eq. \eqref{eq:ncr} and \eqref{eq:acr} are used to calculate the contribution rate of ECG leads towards each diagnosis class, which is utilized for population level interpretation of the deep learning model. Fig. \ref{fig:shap2} demonstrates the contribution rate of ECG leads towards diagnostic classes in the 12-lead deep model. The first observation is that from the average perspective, lead II, aVR, V1, V2, V5, and V6 are the most important leads in the 12-lead model. Secondly, lead V1 and V2 play an important role in LBBB diagnosis with a high contribution rate of around 0.3. For the diagnosis of RBBB, lead V1 and V2 are also important with a high contribution rate. These findings match the diagnostic criteria of LBBB and RBBB \cite{alventosa2019right, goldberger2017clinical}. We also observe some leads (III, aVL) are associated with a low contribution rate which means these leads are possibly neglected in the 12-lead ECG model. This may be because of feature interactions among ECG leads (e.g., lead III is the difference between lead II and lead I).

\section{Conclusion}
In this paper, we developed a deep neural network for multi-label classification of cardiac arrhythmias in 12-lead ECG recordings. The neural network uses stacked residual blocks for feature extraction and a fully connected layer for prediction. Model performance was evaluated on the CPSC2018 dataset and the experiment results empirically showed the effectiveness and efficacy of the proposed network architecture. We compared the performance of the deep learning model with baseline classifiers learned from expert features. The result showed the deep learning model significantly outperformed baseline classifiers. We also trained the model on single-lead ECGs and the result showed lead I, aVR, and V5 are the best-performing leads. Besides, we employed the SHAP method to interpret the model's behavior at both patient level and population level and could help clinical decision making. In the future, we plan to investigate the deep model's ability of generalization to other patients from different ethnicities and robustness to adversarial examples.



\bibliographystyle{IEEEtran}
\bibliography{ref}

\begin{thebibliography}{10}
\providecommand{\url}[1]{#1}
\csname url@samestyle\endcsname
\providecommand{\newblock}{\relax}
\providecommand{\bibinfo}[2]{#2}
\providecommand{\BIBentrySTDinterwordspacing}{\spaceskip=0pt\relax}
\providecommand{\BIBentryALTinterwordstretchfactor}{4}
\providecommand{\BIBentryALTinterwordspacing}{\spaceskip=\fontdimen2\font plus
\BIBentryALTinterwordstretchfactor\fontdimen3\font minus
  \fontdimen4\font\relax}
\providecommand{\BIBforeignlanguage}[2]{{%
\expandafter\ifx\csname l@#1\endcsname\relax
\typeout{** WARNING: IEEEtran.bst: No hyphenation pattern has been}%
\typeout{** loaded for the language `#1'. Using the pattern for}%
\typeout{** the default language instead.}%
\else
\language=\csname l@#1\endcsname
\fi
#2}}
\providecommand{\BIBdecl}{\relax}
\BIBdecl

\bibitem{virani2020heart}
S.~S. Virani, A.~Alonso, E.~J. Benjamin, M.~S. Bittencourt, C.~W. Callaway,
  A.~P. Carson, A.~M. Chamberlain, A.~R. Chang, S.~Cheng, F.~N. Delling
  \emph{et~al.}, ``Heart disease and stroke statistics—2020 update: a report
  from the american heart association,'' \emph{Circulation}, pp. E139--E596,
  2020.

\bibitem{bogun2004misdiagnosis}
F.~Bogun, D.~Anh, G.~Kalahasty, E.~Wissner, C.~B. Serhal, R.~Bazzi, W.~D.
  Weaver, and C.~Schuger, ``Misdiagnosis of atrial fibrillation and its
  clinical consequences,'' \emph{The American journal of medicine}, vol. 117,
  no.~9, pp. 636--642, 2004.

\bibitem{holst1999confident}
H.~Holst, M.~Ohlsson, C.~Peterson, and L.~Edenbrandt, ``A confident decision
  support system for interpreting electrocardiograms,'' \emph{Clinical
  Physiology}, vol.~19, no.~5, pp. 410--418, 1999.

\bibitem{world2014global}
W.~H. Organization \emph{et~al.}, \emph{Global status report on noncommunicable
  diseases 2014}.\hskip 1em plus 0.5em minus 0.4em\relax World Health
  Organization, 2014, no. WHO/NMH/NVI/15.1.

\bibitem{moody2001impact}
G.~B. Moody and R.~G. Mark, ``The impact of the mit-bih arrhythmia database,''
  \emph{IEEE Engineering in Medicine and Biology Magazine}, vol.~20, no.~3, pp.
  45--50, 2001.

\bibitem{clifford2017af}
G.~D. Clifford, C.~Liu, B.~Moody, H.~L. Li-wei, I.~Silva, Q.~Li, A.~Johnson,
  and R.~G. Mark, ``Af classification from a short single lead ecg recording:
  the physionet/computing in cardiology challenge 2017,'' in \emph{2017
  Computing in Cardiology (CinC)}.\hskip 1em plus 0.5em minus 0.4em\relax IEEE,
  2017, pp. 1--4.

\bibitem{liu2018open}
F.~Liu, C.~Liu, L.~Zhao, X.~Zhang, X.~Wu, X.~Xu, Y.~Liu, C.~Ma, S.~Wei, Z.~He
  \emph{et~al.}, ``An open access database for evaluating the algorithms of
  electrocardiogram rhythm and morphology abnormality detection,''
  \emph{Journal of Medical Imaging and Health Informatics}, vol.~8, no.~7, pp.
  1368--1373, 2018.

\bibitem{wagner2020ptb}
P.~Wagner, N.~Strodthoff, R.-D. Bousseljot, D.~Kreiseler, F.~I. Lunze,
  W.~Samek, and T.~Schaeffter, ``Ptb-xl, a large publicly available
  electrocardiography dataset,'' \emph{Scientific Data}, vol.~7, no.~1, pp.
  1--15, 2020.

\bibitem{lundberg2017unified}
S.~M. Lundberg and S.-I. Lee, ``A unified approach to interpreting model
  predictions,'' in \emph{Advances in neural information processing systems},
  2017, pp. 4765--4774.

\bibitem{jambukia2015classification}
S.~H. Jambukia, V.~K. Dabhi, and H.~B. Prajapati, ``Classification of ecg
  signals using machine learning techniques: A survey,'' in \emph{2015
  International Conference on Advances in Computer Engineering and
  Applications}.\hskip 1em plus 0.5em minus 0.4em\relax IEEE, 2015, pp.
  714--721.

\bibitem{macfarlane2005university}
P.~Macfarlane, B.~Devine, and E.~Clark, ``The university of glasgow (uni-g) ecg
  analysis program,'' in \emph{Computers in Cardiology, 2005}.\hskip 1em plus
  0.5em minus 0.4em\relax IEEE, 2005, pp. 451--454.

\bibitem{de2000using}
P.~De~Chazal, B.~Celler, and R.~Reilly, ``Using wavelet coefficients for the
  classification of the electrocardiogram,'' in \emph{Proceedings of the 22nd
  Annual International Conference of the IEEE Engineering in Medicine and
  Biology Society (Cat. No. 00CH37143)}, vol.~1.\hskip 1em plus 0.5em minus
  0.4em\relax IEEE, 2000, pp. 64--67.

\bibitem{datta2017identifying}
S.~Datta, C.~Puri, A.~Mukherjee, R.~Banerjee, A.~D. Choudhury, R.~Singh,
  A.~Ukil, S.~Bandyopadhyay, A.~Pal, and S.~Khandelwal, ``Identifying normal,
  af and other abnormal ecg rhythms using a cascaded binary classifier,'' in
  \emph{2017 Computing in cardiology (cinc)}.\hskip 1em plus 0.5em minus
  0.4em\relax IEEE, 2017, pp. 1--4.

\bibitem{lecun2015deep}
Y.~LeCun, Y.~Bengio, and G.~Hinton, ``Deep learning,'' \emph{Nature}, vol. 521,
  no. 7553, pp. 436--444, 2015.

\bibitem{esteva2019guide}
A.~Esteva, A.~Robicquet, B.~Ramsundar, V.~Kuleshov, M.~DePristo, K.~Chou,
  C.~Cui, G.~Corrado, S.~Thrun, and J.~Dean, ``A guide to deep learning in
  healthcare,'' \emph{Nature medicine}, vol.~25, no.~1, pp. 24--29, 2019.

\bibitem{hong2020opportunities}
S.~Hong, Y.~Zhou, J.~Shang, C.~Xiao, and J.~Sun, ``Opportunities and challenges
  of deep learning methods for electrocardiogram data: A systematic review,''
  \emph{Computers in Biology and Medicine}, p. 103801, 2020.

\bibitem{hannun2019cardiologist}
A.~Y. Hannun, P.~Rajpurkar, M.~Haghpanahi, G.~H. Tison, C.~Bourn, M.~P.
  Turakhia, and A.~Y. Ng, ``Cardiologist-level arrhythmia detection and
  classification in ambulatory electrocardiograms using a deep neural
  network,'' \emph{Nature medicine}, vol.~25, no.~1, p.~65, 2019.

\bibitem{he2019automatic}
R.~He, Y.~Liu, K.~Wang, N.~Zhao, Y.~Yuan, Q.~Li, and H.~Zhang, ``Automatic
  cardiac arrhythmia classification using combination of deep residual network
  and bidirectional lstm,'' \emph{IEEE Access}, vol.~7, pp. 102\,119--102\,135,
  2019.

\bibitem{zhu2020automatic}
H.~Zhu, C.~Cheng, H.~Yin, X.~Li, P.~Zuo, J.~Ding, F.~Lin, J.~Wang, B.~Zhou,
  Y.~Li \emph{et~al.}, ``Automatic multilabel electrocardiogram diagnosis of
  heart rhythm or conduction abnormalities with deep learning: a cohort
  study,'' \emph{The Lancet Digital Health}, 2020.

\bibitem{chen2020detection}
T.-M. Chen, C.-H. Huang, E.~S. Shih, Y.-F. Hu, and M.-J. Hwang, ``Detection and
  classification of cardiac arrhythmias by a challenge-best deep learning
  neural network model,'' \emph{Iscience}, vol.~23, no.~3, p. 100886, 2020.

\bibitem{ullah2020classification}
A.~Ullah, S.~M. Anwar, M.~Bilal, and R.~M. Mehmood, ``Classification of
  arrhythmia by using deep learning with 2-d ecg spectral image
  representation,'' \emph{Remote Sensing}, vol.~12, no.~10, p. 1685, 2020.

\bibitem{hong2017encase}
S.~Hong, M.~Wu, Y.~Zhou, Q.~Wang, J.~Shang, H.~Li, and J.~Xie, ``Encase: An
  ensemble classifier for ecg classification using expert features and deep
  neural networks,'' in \emph{2017 Computing in cardiology (cinc)}.\hskip 1em
  plus 0.5em minus 0.4em\relax IEEE, 2017, pp. 1--4.

\bibitem{liu2018automatic}
Z.~Liu, X.~Meng, J.~Cui, Z.~Huang, and J.~Wu, ``Automatic identification of
  abnormalities in 12-lead ecgs using expert features and convolutional neural
  networks,'' in \emph{2018 International Conference on Sensor Networks and
  Signal Processing (SNSP)}.\hskip 1em plus 0.5em minus 0.4em\relax IEEE, 2018,
  pp. 163--167.

\bibitem{he2016deep}
K.~He, X.~Zhang, S.~Ren, and J.~Sun, ``Deep residual learning for image
  recognition,'' in \emph{Proceedings of the IEEE conference on computer vision
  and pattern recognition}, 2016, pp. 770--778.

\bibitem{lundberg2020local}
S.~M. Lundberg, G.~Erion, H.~Chen, A.~DeGrave, J.~M. Prutkin, B.~Nair, R.~Katz,
  J.~Himmelfarb, N.~Bansal, and S.-I. Lee, ``From local explanations to global
  understanding with explainable ai for trees,'' \emph{Nature machine
  intelligence}, vol.~2, no.~1, pp. 2522--5839, 2020.

\bibitem{li2020time}
X.~Li, X.~Xu, F.~Xie, X.~Xu, Y.~Sun, X.~Liu, X.~Jia, Y.~Kang, L.~Xie, F.~Wang
  \emph{et~al.}, ``A time-phased machine learning model for real-time
  prediction of sepsis in critical care,'' \emph{Critical Care Medicine}, 2020.

\bibitem{paszke2019pytorch}
A.~Paszke, S.~Gross, F.~Massa, A.~Lerer, J.~Bradbury, G.~Chanan, T.~Killeen,
  Z.~Lin, N.~Gimelshein, L.~Antiga \emph{et~al.}, ``Pytorch: An imperative
  style, high-performance deep learning library,'' in \emph{Advances in Neural
  Information Processing Systems}, 2019, pp. 8024--8035.

\bibitem{jlb2015adam}
D.~P.~K. JLB, ``Adam: A method for stochastic optimization,'' in \emph{3rd
  international conference for learning representations, San Diego}, 2015.

\bibitem{mccabe2013physician}
J.~M. McCabe, E.~J. Armstrong, I.~Ku, A.~Kulkarni, K.~S. Hoffmayer, P.~D.
  Bhave, S.~W. Waldo, P.~Hsue, J.~C. Stein, G.~M. Marcus \emph{et~al.},
  ``Physician accuracy in interpreting potential st-segment elevation
  myocardial infarction electrocardiograms,'' \emph{Journal of the American
  Heart Association}, vol.~2, no.~5, p. e000268, 2013.

\bibitem{pedregosa2011scikit}
F.~Pedregosa, G.~Varoquaux, A.~Gramfort, V.~Michel, B.~Thirion, O.~Grisel,
  M.~Blondel, P.~Prettenhofer, R.~Weiss, V.~Dubourg \emph{et~al.},
  ``Scikit-learn: Machine learning in python,'' \emph{Journal of machine
  learning research}, vol.~12, no. Oct, pp. 2825--2830, 2011.

\bibitem{ke2017lightgbm}
G.~Ke, Q.~Meng, T.~Finley, T.~Wang, W.~Chen, W.~Ma, Q.~Ye, and T.-Y. Liu,
  ``Lightgbm: A highly efficient gradient boosting decision tree,'' in
  \emph{Advances in neural information processing systems}, 2017, pp.
  3146--3154.

\bibitem{apple4ecg}
\BIBentryALTinterwordspacing
Apple. Taking an ecg with the ecg app on apple watch series 4 or later.
  [Online]. Available: \url{https://support.apple.com/en-us/HT208955}
\BIBentrySTDinterwordspacing

\bibitem{gutierrez2011atrial}
C.~Gutierrez and D.~G. Blanchard, ``Atrial fibrillation: diagnosis and
  treatment,'' \emph{American family physician}, vol.~83, no.~1, pp. 61--68,
  2011.

\bibitem{barold2006first}
S.~S. Barold, A.~Ilercil, F.~Leonelli, and B.~Herweg, ``First-degree
  atrioventricular block,'' \emph{Journal of Interventional Cardiac
  Electrophysiology}, vol.~17, no.~2, pp. 139--152, 2006.

\bibitem{alventosa2019right}
M.~Alventosa-Zaidin, L.~Guix~Font, M.~Benitez~Camps, C.~Roca~Saumell, G.~Pera,
  M.~T. Alzamora~Sas, R.~For{\'e}s~Raurell, O.~Rebagliato~Nadal,
  A.~Dalf{\'o}-Baqu{\'e}, and J.~Brugada~Terradellas, ``Right bundle branch
  block: Prevalence, incidence, and cardiovascular morbidity and mortality in
  the general population,'' \emph{European Journal of General Practice},
  vol.~25, no.~3, pp. 109--115, 2019.

\bibitem{goldberger2017clinical}
A.~L. Goldberger, Z.~D. Goldberger, and A.~Shvilkin, \emph{Clinical
  electrocardiography: a simplified approach e-book}.\hskip 1em plus 0.5em
  minus 0.4em\relax Elsevier Health Sciences, 2017.

\end{thebibliography}

\end{document}